\documentclass{article}

\usepackage{arxiv}

\usepackage[utf8]{inputenc} % allow utf-8 input
\usepackage[T1]{fontenc}    % use 8-bit T1 fonts
\usepackage{hyperref}       % hyperlinks
\usepackage{url}            % simple URL typesetting
\usepackage{booktabs}       % professional-quality tables
\usepackage{amsfonts}       % blackboard math symbols
\usepackage{nicefrac}       % compact symbols for 1/2, etc.
\usepackage{microtype}      % microtypography
\usepackage{lipsum}		% Can be removed after putting your text content
\usepackage{graphicx}
\usepackage{natbib}
\usepackage{doi}
\usepackage{accessibility}

\title{Evaluating the Role of Large Language Models in Legal Practice in India}

\date{January 15, 2025}	% Here you can change the date presented in the paper title
\author{ \href{https://orcid.org/0000-0002-9625-4106}{\includegraphics[scale=0.06]{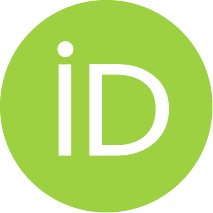}\hspace{1mm}
Rahul Hemrajani} \\
    Assistant Professor of Law\\
	Faculty Director, JSW Centre for the Future of Law\\
	National Law School of India\\
	Bengaluru, India \\
	\texttt{rahul.hemrajani@nls.ac.in} \\}

\renewcommand{\headeright}{}
\renewcommand{\undertitle}{}
\renewcommand{\shorttitle}{Evaluating the Role of Large Language Models in Legal Practice in India}

\hypersetup{
pdftitle={Evaluating the Role of Large Language Models in Legal Practice in India},
pdfsubject={Artificial Intelligence and the Law},
pdfauthor={Rahul Hemrajani},
pdfkeywords={Artificial Intelligence (AI); AI and the Law; Large Language Models (LLMs)},
}

\begin{document}
\maketitle

\begin{abstract}
The integration of Artificial Intelligence (AI) into the legal profession raises significant questions about the capacity of Large Language Models (LLMs) to perform key legal tasks. In this paper, I empirically evaluate how well LLMs, such as GPT-4, Claude, and ChatGPT, perform key legal tasks in the Indian context, including issue spotting, legal drafting, advice, research, and reasoning. Through a survey experiment, I compare outputs from LLMs with those of a junior lawyer, with advanced law students rating the work on helpfulness, accuracy, and comprehensiveness. LLMs excel in drafting and issue spotting, often matching or surpassing human work. However, they struggle with specialised legal research, frequently generating hallucinations—factually incorrect or fabricated outputs.  I conclude that while LLMs can augment certain legal tasks, human expertise remains essential for nuanced reasoning and the precise application of law.
\end{abstract}

% keywords can be removed

\section{Introduction}

The integration of Artificial Intelligence (AI) into various professional fields has ignited debates about the future of work. The legal profession is no exception. With the advent of Large Language Models, AI tools have grown increasingly capable of undertaking tasks traditionally performed by lawyers. While some scholars predict that AI will revolutionise the legal profession by automating a wide range of tasks, others argue that AI's role will be limited to handling repetitive, low-value tasks, such as document review or transcription. This rapid technological development raises a pressing question: how and to what extent can generative AI replace humans for legal work?

AI is not a one-size-fits-all solution, and its effectiveness varies significantly across different types of legal tasks. The challenge lies in disentangling which aspects of legal practice AI tools can reliably handle and which remain firmly in the human domain. Yet, there has been no systematic, empirical study that fully investigates AI's capabilities and limitations in the legal field, especially in varied jurisdictions. This gap is particularly pronounced in India, where the scarcity of publicly accessible legal data poses unique challenges. Large Language Models (LLMs) like GPT-4 and other AI systems are often trained predominantly on datasets that reflect legal materials from jurisdictions with large amounts of public legal data, such as the United States or the European Union. Consequently, their effectiveness in the Indian legal context—where case law, statutes, and practice materials are not as readily available—is an open question.

In this paper, I empirically evaluate the ability of LLMs to perform various legal tasks in an Indian law context. Specifically, I test how different LLM models perform on five key legal tasks: issue spotting, legal drafting, providing legal advice, conducting legal research, and demonstrating legal reasoning. I find that that LLMs generally perform as well as, or better than, human experts in several language-based legal tasks, such as drafting legal documents or identifying issues within a given fact pattern. However, I find that LLMs struggle significantly with tasks involving specialized legal research.

I make three main contributions:
\begin{enumerate}
    \item \textbf{First empirical analysis of LLMs in Indian legal tasks:} I present the first systematic study evaluating the performance of Large Language Models (LLMs) in key legal tasks within the Indian legal context, addressing the unique challenges posed by limited digitised and publicly accessible legal data in India.
    
    \item \textbf{Multiple legal Task and model-specific evaluation:} I conduct a comparative analysis of multiple LLMs, including GPT-4, Claude, ChatGPT 3.5, Gemini, and Llama 2, across five critical legal tasks—issue spotting, legal drafting, legal advice, legal research, and legal reasoning—highlighting strengths and limitations specific to each model and task.
    
    \item \textbf{Qualitative insights into lawyer preferences:} I collect and analyse qualitative feedback from advanced law students, revealing why certain LLM outputs are preferred over others. This provides actionable insights into the factors that influence the perceived quality of AI-generated legal outputs, such as clarity, relevance, and professionalism.
\end{enumerate}

\section{Background and Related Literature}
The development of Large Language Models (LLMs) have emerged as a transformative force across various industries, reshaping how businesses operate and innovate. From healthcare to finance, these AI-driven tools are disrupting traditional practices by automating complex tasks, generating insights from vast amounts of data, and providing decision-making support. The legal industry is no exception to this wave of disruption \citep{Susskind2023}. In the last 3 years, several law firms and companies have started using AI tools to automate legal processes, challenging the conventional roles of legal professionals and prompting a revaluation of how legal services are delivered \citep{Bhavani2022}. In fact, Goldman Sachs has projected that up to 44\% of legal tasks might be susceptible to automation by AI \citep{Briggs2023}

Lawyers across the world are starting to recognise this change. A survey of lawyers in the US found that 46\% of lawyers already use AI for work \citep{Pacheco2024},  while another found that 73\% of lawyers expect to integrate GenAI in their legal work soon \citep{WoltersKluwer2023}. Similarly, out of eight practicing lawyers in India interviewed for this study, three said that they “use AI regularly for legal work”, three have “experimented with GenAI tools for legal work” and only two have “never used AI for their legal work”.\footnote{The lawyers were interviewed through a semi-structured questionnaire. Four of these lawyers are litigators practicing in various courts in India and four are part of law firms.}  One lawyer, who is an Advocate-on-Record in the Supreme Court of India said that they regularly use ChatGPT for generating first draft of contracts and specific contractual clauses. Another, who works in a top-firm in India stated that, “I use AI in a a lot of our regular drafting and research work.”   One lawyer even predicted that AI might replace human lawyers soon stating, “Our firm is engaging in a top-to-bottom integration of AI into our practice. Soon I expect AI will be able to do all the work that would have been done by A0s (Junior Associates with less than one year of experience).”

There are three reasons why LLMs (Large Language Models) are particularly well-suited for the law. First, legal work is fundamentally language-intensive, involving tasks such as drafting contracts, summarizing case law, and interpreting statutes—areas where LLMs excel due to their sophisticated language generation and comprehension abilities. Second, most legal tasks require a vast and up-to-date specialised knowledge base. LLMs, trained on extensive datasets that include publicly available case laws, statutes, regulations, and scholarly articles, can provide immediate access to this information faster than traditional methods. Third, legal work often depends on analogical reasoning—drawing parallels between different cases or legal principles to make arguments or predictions. LLMs, by their very nature, are designed to detect patterns and principles within vast bodies of text, making them adept at finding analogous cases, identifying core legal principles, and applying them to new scenarios.

Many studies have demonstrated that LLM tools, such as ChatGPT, can competently pass law school and professional qualification exams for the law. For instance, one study found that ChatGPT-3.5 achieved an average grade of C+ on law school exams, sufficient to pass \citep{Choi2021}. Similarly, LLMs have shown the capability to pass professional qualifying examinations for the legal field. A study revealed that GPT-3.5 achieved a 50.3\% correct rate on a full NCBE Multistate Bar Exam (MBE) practice test, a key exam for qualifying to practice law in the United States \citep{Bommarito2022}. Building on this, GPT-4 demonstrated even stronger performance on the MBE, significantly outperforming both human test-takers and previous models, with a 26\% improvement over ChatGPT-3.5 and outperforming humans in five of seven subject areas (Katz et al. 2024). Likewise, another study found that GPT-3.5 scored 58.7\% and GPT-4 scored 75\% on the All India Bar Examination—an exam with a pass rate of less than 50\% for human candidates—indicating LLMs' potential to pass rigorous legal certification tests \citep{Karn2023}.

However, these results do not account for inconsistencies in LLM performance across different contexts. For example, one study found that while ChatGPT performed well in law school subjects such as Jurisprudence and International Tax Law, which require generic knowledge, it received an F grade in Employment Law and Company Law in Hong Kong, where the application of localized legal knowledge is crucial \citep{Hargreaves2023}. Another study critiqued the findings of the GPT-4 MBE study, highlighting that GPT-4's performance varied significantly when evaluated against different groups of test-takers, with its performance dropping to the 15th percentile on essay questions \citep{Martinez2024}. Similarly, a study on the Brazilian law examination found that GPT-4 performed well on multiple-choice questions but struggled significantly on essay-type questions \citep{Freitas2023}. These findings align with another study that compared GPT-4's exam performance with human students, revealing that its results varied widely depending on the quality and specificity of the prompts used \citep{Choi2023}.

Empirical evidence on the effectiveness of AI for real-world legal tasks also remain scant. Many studies have highlighted the challenges associated with using AI models for legal work, noting that while GPT models can provide a good interactive experience, they sometimes generate "hallucinations" or inaccurate responses \citep{Tan2023,Savelka2023}. The few empirical assessments of the potential of using AI in specific legal tasks have seen mixed results \citep{Chalkidis2023}. For instance, one study tested GPT-4 and Mixtral8x7B on their ability to answer legal questions and found that while LLMs can generate sophisticated responses, there was a noticeable preference for human answers due to their clarity and directness \citep{Bhambhoria2024}. Similarly, another study examined ChatGPT and evaluated its performance across 16 different criteria as rated by humans. The conclusions noted that AI could be helpful in some instances but was generally limited in its utility, reflecting an ongoing gap between AI and human experts in legal comprehension and expression \citep{Hagan2023}. On the other hand, a more recent study compared various LLMs with junior lawyers on tasks such as contract reviews. The findings revealed that AI models performed comparably to, or even better than, junior lawyers in specific scenarios, particularly when the task was well-defined and involved repetitive textual analysis \citep{Martin2024}.

Despite this scholarship, lawyers still lack clear guidance on which specific legal tasks are suitable for AI assistance and which AI tools are best suited for these tasks. First, most studies focus on evaluating AI performance on individual legal tasks, such as answering legal questions, reviewing contracts, or drafting legal documents, without offering a comparison of the utility of AI across different types of tasks. Second, there is a lack of studies that compare the performance of different LLM models, such as GPT, Claude, and Gemini, on the same legal tasks under similar conditions. Third, most of the existing studies are conducted in developed countries where access to extensive public legal data supports the training and accuracy of LLMs. This raises concerns about the applicability and performance of these models in developing countries, like India, where the legal context is highly localized and due to the lack of public data, may not have been part of the training corpus of LLMs. 

This paper then aims to explore a broad range of tasks that are integral to the legal profession in India and examine where and how can AI assist human-lawyers to offer better and more effective legal services. To the best of my knowledge, it is first empirical study to systematically evaluate and compare how different LLMs perform across various legal tasks specific to the needs and practices of Indian lawyers.

\section{Research Methods}

As part of this study, I designed a mixed-method approach that combines a survey experiment with qualitative interviews to assess the performance of Large Language Models (LLMs) in various legal tasks. The participants were 50 advanced law students.\footnote{The students were students of 3rd,4th and 5th years of the BA LLB program at [blinded] as well as 2nd year students of the LLB program.  } This included students enrolled in an elective course on AI and the law, for whom participation in the study was a course requirement as well as other student volunteers. No student was paid for their participation.

The study utilized a consumer law problem as the basis for the legal tasks, specifically involving a scenario where an individual discovers an insect in her drink while dining at a restaurant. Consumer law represents a bounded and relatively limited body of case law. This makes it an ideal domain for evaluating AI-generated legal content since the legal principles are well-established and the factual matrix is common to civil legal disputes. Consumer law also involves clear standards for liability and remedies, which allows for consistent and objective evaluation of outputs across different legal tasks, ensuring that both the strengths and limitations of AI models can be effectively assessed.

To evaluate the capabilities of LLMs, outputs were generated for five distinct legal tasks: issue spotting, legal drafting, providing legal advice, conducting legal research, and demonstrating legal reasoning. Outputs for these tasks were produced by six different “participants”: five LLMs—ChatGPT 3.5, Claude 3, GPT-4, Gemini, and Llama 2 (through Poe)—and a human expert. These were chosen for their high ranking on the LLMYS leaderboard in April 2024 and to represent a diverse range of capabilities and access options. ChatGPT 3.5 is a free model, while Claude 3, GPT-4, and Gemini are paid, advanced models, offering varying levels of sophistication. Gemini stood out for its unique ability to access the internet, potentially providing more current and context-aware responses. Llama 2, an open-source model, allows for flexibility and adaptability in research settings. This human expert, a recently qualified lawyer with one year of experience in litigation, including consumer law, was included to serve as a baseline for comparison against the AI models.\footnote{This level of experience provided a competent, practical benchmark for comparison, as the study aimed to evaluate AI models against a capable, though not highly experienced or specialised, human standard. The expert was given the same prompt as the LLM chatbots, and was given 6 hours to complete each task. The expert was given access to all resources at the [blinded] but was instructed to not use any AI or LLM tool. The expert was paid for their time and effort and were not told about what their outputs would be used for. } 

Each participant, human or AI, was given identical prompts developed through careful prompt engineering to ensure that the outputs would be as relevant and complete as possible for each task.  We engaged three research assistants to try out various prompt strategies for different legal tasks using the guides provided by the AI companies. Each research assistant was instructed to evaluate and rate the quality of outputs generated by the LLM based on their prompt inputs. The purpose of this exercise was to identify effective prompting techniques specifically for legal tasks. Based on this, we finalised a final prompt that was used in a one-shot manner for all AI outputs.

All generated outputs were anonymised and presented in a randomized order to the student raters, who were unaware that one set of responses was created by a human expert. The students were asked to evaluate the outputs based on three criteria: helpfulness, accuracy, and comprehensiveness, using a five-point Likert scale where 1 indicated "very poor" and 5 indicated "excellent." These criteria were selected to capture different dimensions of quality in legal work: helpfulness relates to the utility of the information provided, accuracy pertains to factual correctness and adherence to legal standards, and comprehensiveness reflects the thoroughness and depth of the analysis or advice offered. Students were also asked to speculate whether each output was generated by a human or an AI.  

Beyond these quantitative evaluations, the survey also gathered qualitative feedback from the students. After rating each output, participants were required to provide open-ended responses explaining their scores and reasoning for their judgment about whether the text was AI or human-generated. This qualitative component allowed us to explore the reasoning behind perceived strengths and weaknesses, such as why a particular output was deemed comprehensive or why another was rated low for accuracy. The full survey took the average respondent 2 hours 15 minutes to complete.

\section{Results}

\subsection{Issue-Spotting and Legal Text Summarisation}

Issue-spotting in the legal profession refers to the ability of a lawyer or legal professional to identify the key legal issues, facts, and questions that arise from a set of circumstances or documents. Information extraction and classification have been long-standing focus areas for automation in the legal industry \citep{Hachey2006}. Law firms conducting due diligence in mergers and acquisitions, or litigators sifting through vast records in appellate cases, rely heavily on their ability to quickly identify, retrieve and tag relevant information from large volumes of documents. Even before the rise of Generative AI, there was considerable progress in developing tools for these purposes. Law firms and legal departments invested in proprietary software solutions designed to automatically summarize, tag, and categorize documents \citep{Jung2019,Armour2020}. These tools, often equipped with natural language processing and machine learning algorithms, helped in reducing manual workload, minimizing human error, and enhancing the speed of document review and analysis \citep{Ashley2018}.

In the specific task of issue-spotting, some studies have found that AI particularly when pre-trained on specific types of legal documents, can outperform human experts in identifying key legal issues, especially in repetitive and rule-based contexts \citep{LawGeex2018}. LLMs have also demonstrated robust capabilities in tasks like semantic annotation and rule classification \citep{Medvedeva2023,Jang2024}. One empirical study found that some AI tools such as ChatGPT 4, were able to perform better than junior lawyers in contract review and issue-spotting tasks \citep{Martin2024}. 

For the issue-spotting task, participants were given a legal scenario involving a consumer grievance against a hotel in India, presented in a structured legal format. The case details a complaint filed by a consumer who allegedly found insects in a juice served at a hotel, suffered subsequent illness, and sought compensation for her distress and expenses. The documents provided included a detailed complaint submitted to the District Consumer Disputes Redressal Commission by the complainant, outlining the facts, reliefs sought, and legal grounds under the Consumer Protection Act, 2019, along with a written statement of objections from the opposite party (the hotel), denying the allegations and presenting their defence. Based on these documents, both LLMs and a human expert were tasked with creating a one-page material summary of the case. This summary needed to capture the key elements such as the overall case summary, issues in dispute, evidence presented by both parties, the relief claimed, and whether the complaint met the jurisdictional and limitation criteria. Respondents were then asked to evaluate these summaries. 

\begin{figure}[h]
  \centering
  \includegraphics[width=\linewidth]{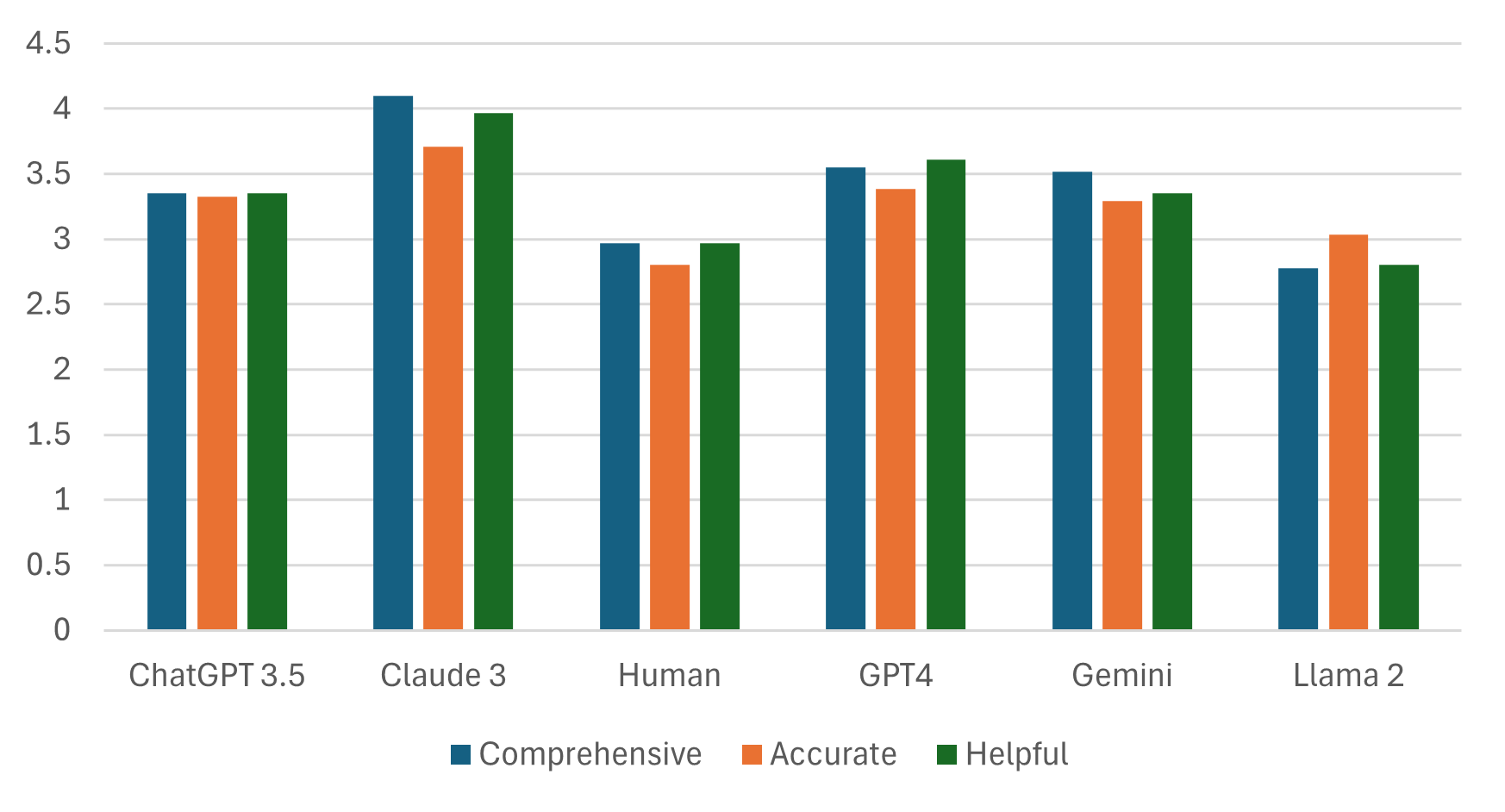}
  \caption{Bar chart of mean evaluation score of outputs for the issue-spotting task.}
  \label{fig:issue-spotting}
\end{figure}

The results of these evaluations are given in Figure \ref{fig:issue-spotting}. Claude 3 emerged as the top performer, receiving the highest average ratings. This suggests that Claude 3 not only provides thorough coverage of legal issues but also does so with a high degree of relevance and usefulness. GPT 4 had comparable, moderately strong performances. ChatGPT 3.5, the free version available to the public had the lowest ratings among the LLMs. The human expert performed poorly, and was rated slightly lower than all other LLMs.

Table \ref{tab:human_responses} shows the assessment of whether participants considered the output to be human or AI-generated. Claude 3’s outputs were frequently mistaken for human-generated content by the student evaluators, highlighting its ability to produce summaries with human-like quality and style. Conversely, models like GPT 4 and Llama 2 were more readily identified as AI-generated.

\begin{table}[ht]
\centering
\caption{\% Respondents who identified the response as "human" across models and tasks.}
\label{tab:human_responses}
\resizebox{\linewidth}{!}{%
  \begin{tabular}{lccccc}
    \toprule
    \textbf{Model} & \textbf{Summary} & \textbf{Advice} & \textbf{Drafting} & \textbf{Research} & \textbf{Reasoning} \\
    \midrule
    ChatGPT 3.5 & 27\% & 33\% & 33\% & 6\%  & 24\% \\
    Claude 3    & 61\% & 27\% & 61\% & 33\% & 52\% \\
    Human       & 27\% & 48\% & 61\% & 36\% & 24\% \\
    GPT4        & 24\% & 55\% & 24\% & 9\%  & 18\% \\
    Gemini      & 42\% & 24\% & 33\% & 6\%  & 27\% \\
    Llama 2     & 12\% & 21\% & 27\% & 12\% & 30\% \\
    \bottomrule
  \end{tabular}%
}
\end{table}

The qualitative responses provide more specific insights into these scores. GPT-4 were described as "comprehensively mentioning the factual part" and having a "clear and structured outline of the case." Evaluators noted that GPT-4 "highlights the issues quite comprehensively" and "provides a good summary of the reliefs sought and the applicability of the limitation period." Claude 3, was similarly praised for its ability to produce "very well drafted with great clarity" summaries that struck a balance between detail and readability. One evaluator noted, "The response captures all important details in a concise manner which makes it easy and less time-consuming to read." However, even with its strengths, some responses highlighted that Claude 3 "missed the nuances of the original para," such as finer points of the complaint or the rebuttal by the opposite party. Llama 2, which scored the lowest among the models, was frequently criticized for its inaccuracies and "hallucinations"—instances where it fabricated or misrepresented facts. As one evaluator pointed out, "It hallucinates evidence (such as evidence of grievance registration on the hotel’s website)," and another added that it "fails to identify all issues," particularly when they are critical to the case's context. Due to these glaring inaccuracies, Llama 2’s outputs were often easily identified as AI-generated, with one evaluator commenting, "No human would (hopefully) think this is a rational way to summarize—there is so much repetition and nonsense littered." 

Human-generated summaries, which surprisingly received the lowest scores across all models, were also subject to mixed feedback. While they were praised for their ability to “frame issues in the order of inquiry provided by the statute” and “capture the gist of facts from both sides,” evaluators were often disappointed by their lack of detail and thoroughness. Comments such as “too brief and missing essential evidence” and “fails to mention relevant legal provisions” highlight that some human outputs did not meet the standards of detailing expected of them. This led to some misidentifications where evaluators mistook the human outputs for AI, remarking that they “read like LLM summaries lacking in-depth analysis”.

There could be two reasons for these unexpectedly low evaluations of human outputs. First, the human expert might have been aware of the simulated environment of the task and thus delivered more concise summaries, knowing they were for academic exercise rather than real legal use. However, it is worth noting that these summaries were verified by another student and a research assistant for use as training data for another project, which implies some level of review and approval of their quality. Second, participants may have been primed by seeing the more detailed outputs generated by advanced LLMs earlier in the exercise, which could have raised their expectations for what constitutes a thorough summary. As a result, the more concise human summaries may have been judged more harshly against these expectations.

The qualitative feedback and quantitative results together suggest that AI models like Claude 3 and GPT-4 are already performing at a level that rivals, and in some cases surpasses, human outputs in structured tasks like legal summarization. These models excel in generating well-organized, clear, and comprehensive summaries that are better than those produced by human experts. 

\subsection{Legal Drafting}

Legal drafting refers to the skill of translating legal principles and facts into formal, structured texts such as contracts, pleadings, wills, or regulations. A few studies have examined the effectiveness of LLM tools for the purpose of legal drafting. One study, for example, found that AI had advanced drafting skills and “elaborate and enhance the contents based on the simple facts” and demonstrate “the ability to understand simple facts and articulate the legal basis of the claim” \citep{Iu2023}.  In another study, researchers investigated the ability of GPT models to draft complaints in cryptocurrency securities class actions. The study found that AI-generated complaints were nearly as convincing as those written by lawyers, although the AI versions were generally more concise and less detailed \citep{Martin2024}. However, one study has found that LLMs, especially while crafting contracts, make formulaic drafts, which have “a number of fallacies, either because they are incomplete, or because they include clauses that are redundant or inapplicable in a given legal system.” \citep{Giampieri2024}.

For the drafting task, I aimed to evaluate the proficiency of Large Language Models (LLMs) in comparison to human experts in the legal drafting of consumer complaints. Both LLMs and a human expert were instructed to draft a detailed consumer complaint for a client who had suffered food poisoning from an insect found in a drink served by a restaurant. I provided a specific template found in the Consumer Protection Act, 2019 to guide the drafting, with the instruction to avoid using points or headings and to format the content strictly in paragraphs. The complaint was to be as detailed as possible, incorporating the identification of parties, a statement of facts, legal grounds for the complaint, and the relief sought.
 
\begin{figure}[h]
  \centering
  \includegraphics[width=\linewidth]{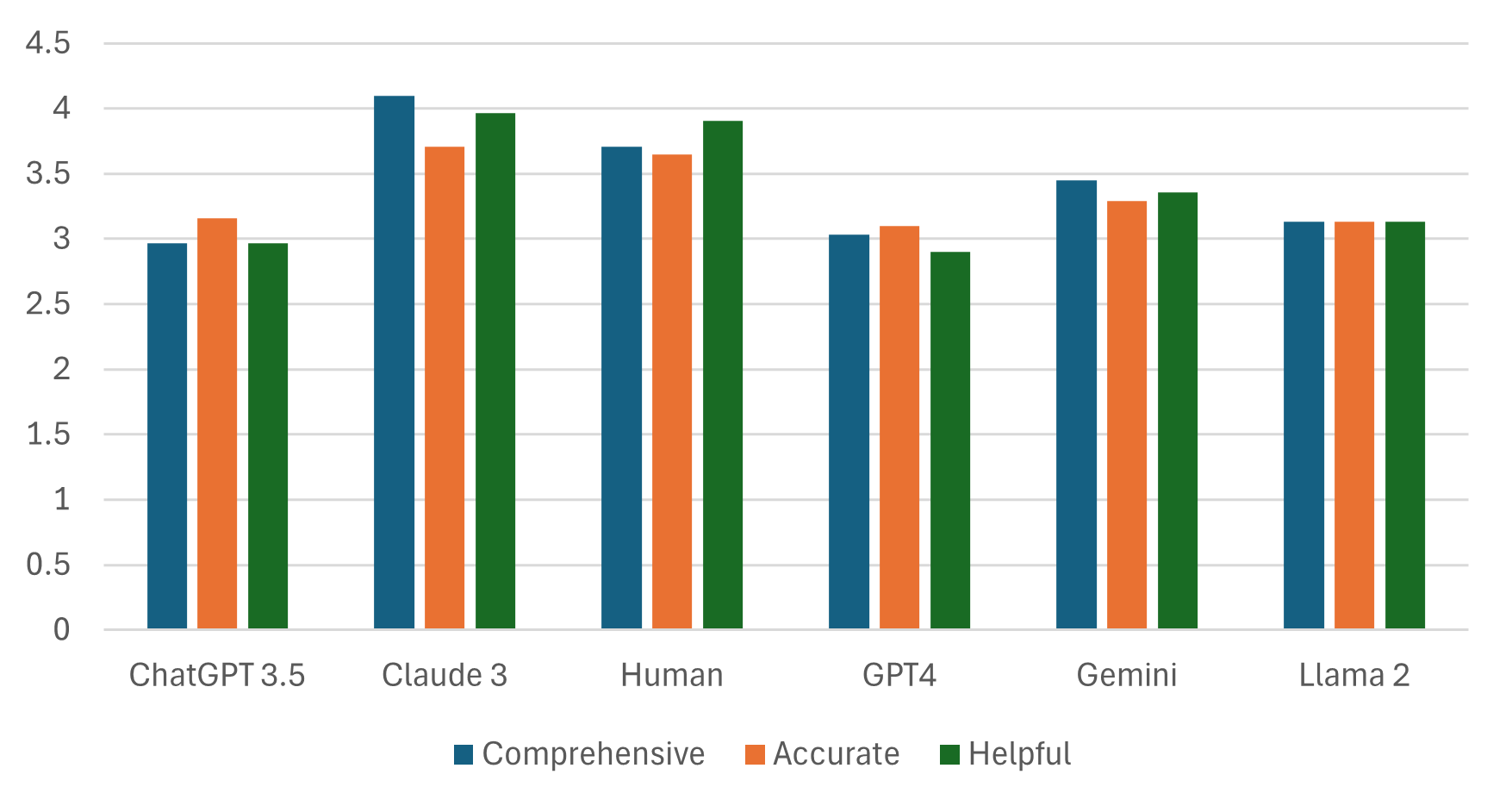}
  \caption{Bar chart of mean evaluation score of outputs for the legal drafting task.}
  \label{fig:drafting}
\end{figure}

Figure \ref{fig:drafting} shows the results for the drafting task. Claude 3 exhibited the highest average scores, indicating a strong ability to generate detailed and precise outputs. Human outputs were rated closely behind Claude 3 in comprehensiveness and accuracy, with average scores around 3.9, demonstrating their ability to maintain a high standard of quality and reliability. Models such as ChatGPT 3.5, Gemini, and Llama 2 consistently received ratings near 3.2 to 3.6 across all criteria, reflecting a satisfactory but moderate level of performance. These models demonstrated the ability to draft coherent outputs but fell short of the higher quality seen in Claude 3 and human outputs. As seen in \ref{tab:human_responses}, both human and claude-3 outputs were most frequently and equally likely to be seen as human. Llama 2 had the lowest count of human identifications, further highlighting its weakness in producing nuanced, human-like text. 

The qualitative feedback indicated Claude consistently received high ratings for its ability to generate legal drafts that are well-structured and organized in a manner that closely resembles human writing. Many evaluators noted that Claude’s responses were "well-formatted" and provided "clear and logical narratives," which made them both easy to read and effective in conveying the core arguments. This model’s ability to incorporate relevant legal provisions, cite appropriate laws, and outline the necessary factual and procedural grounds contributed to its high performance. However, while Claude excelled in adhering to the formal requirements of legal drafting, it sometimes lacked the deep contextual understanding and emotional nuance that human experts brought to the table. For example, Claude's use of emotional language and specific relief justifications sometimes appeared overly formulaic, highlighting an area where it still trails behind human experts in terms of strategic variability and depth.

Conversely, LLaMA was LLaMA's outputs were "rigid and mechanical," with a tendency to strictly adhere to templates without adequately tailoring responses to the specific facts of the case. For instance, LLaMA was criticized for "missing crucial headings" and "incorrectly identifying jurisdictional clauses," demonstrating a weaker grasp of integrating legal reasoning with the facts provided. 

Human experts were consistently rated high for their ability to produce comprehensive, accurate, and highly persuasive legal drafts. Evaluators frequently praised human-generated responses for their "strategic framing" and "attention to detail," which are crucial in legal drafting. Human drafters demonstrated an ability to integrate facts seamlessly with applicable legal principles, craft compelling narratives, and anticipate potential defences from the opposing party. For example, human-generated drafts often included specific witness testimonies, medical records, or detailed breakdowns of damages, which not only added credibility but also significantly strengthened the complainant’s case. This level of contextual understanding, coupled with persuasive language tailored to the specific legal issue, is an area where AI models, including Claude, still lag behind. It would seem then that although AI has made significant strides, particularly models like Claude, the sophistication required for high-stakes legal drafting that involves intricate argumentation and emotional nuance is still best achieved by human drafters. 

\subsection{Legal Advice}

Legal advice involves providing a client with informed guidance on their legal rights, obligations, and potential courses of action based on an analysis of the law and the specific facts of their situation.  When it comes to providing legal advice, previous studies show mixed results. One study found that ChatGPT 3.5 excels in user interaction and offers “an outstanding interactive experience with minimal learning costs for users, allowing them to describe their legal matters using fragmented language and subsequently correct or reinforce the facts during the conversation” \citep{Tan2023}. However, it also hallucinates on occasion. On the other hand, another study found that legal advice given by LLM tools can be ridden with errors \citep{Ryan2024}.

For the legal advice task, respondents were presented with a scenario that required providing detailed legal guidance to a client who had experienced a consumer grievance. The task involved crafting a response email from the perspective of a law firm specializing in consumer law in India. The respondents—both Large Language Models (LLMs) and a human expert—were expected to advise the client on various remedies such as communicating directly with the opposite party, issuing a legal notice, contacting a consumer hotline, and ultimately filing a case "edaakhil", the e-filling portal for consumer dispute resolution. The responses needed to be written in professional, empathetic paragraphs, without the use of bullet points or headings, and had to include comprehensive procedural steps specific to the issue at hand.

\begin{figure}[h]
  \centering
  \includegraphics[width=\linewidth]{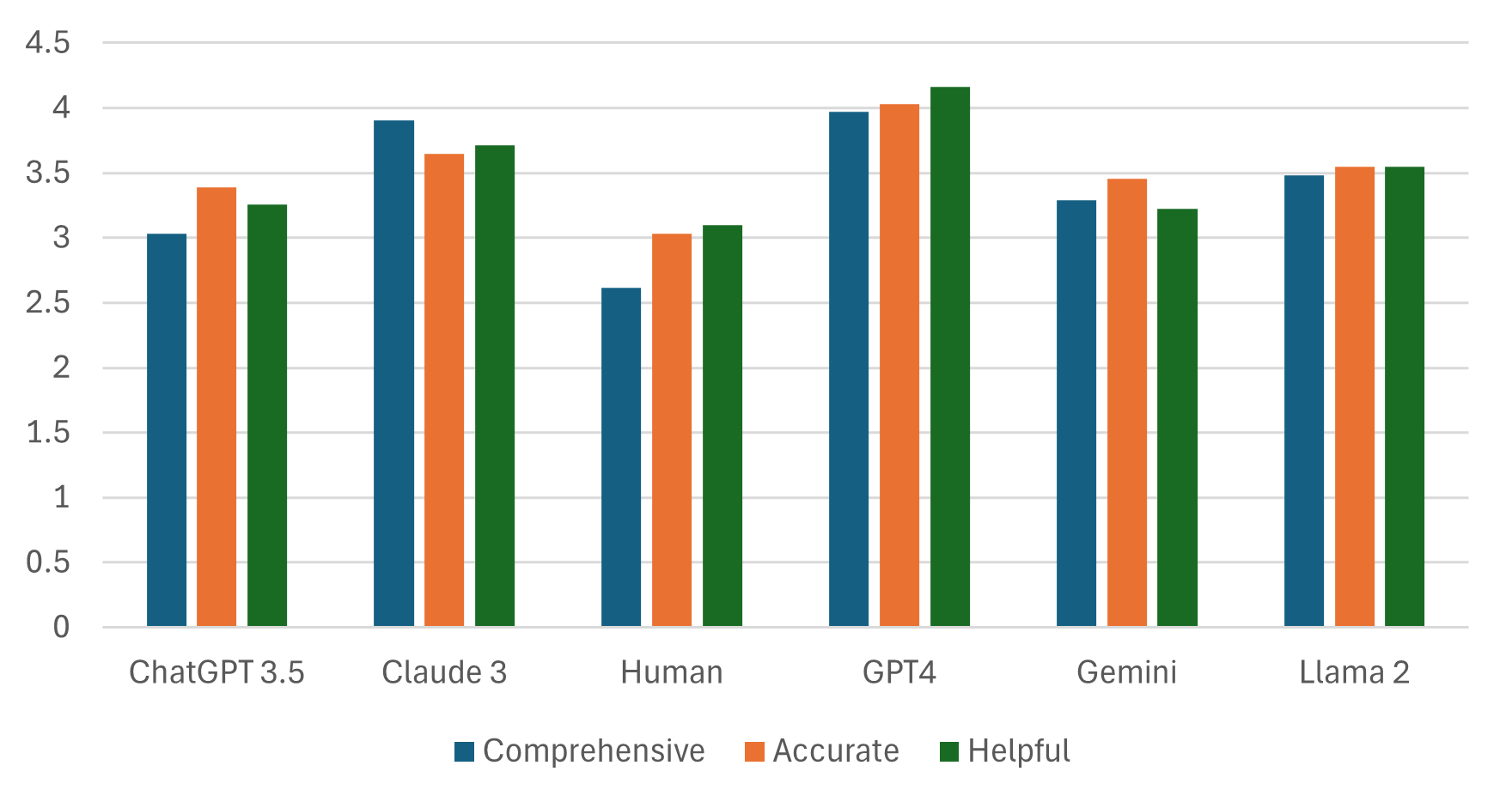}
  \caption{Bar chart of mean evaluation score of outputs for the legal advice task.}
  \label{fig:advice}
\end{figure}

Figure \ref{fig:advice} presents the average scores for the legal advice task. As can be seen, GPT-4 outperformed all other outputs across all three evaluation criteria. Claude 3 and Llama 2 also demonstrated strong performance, particularly in accuracy and comprehensiveness while ChatGPT 3.5 and Gemini received the lowest scores among LLMs. Notably, the human expert scored lower than all AI models. As \ref{tab:human_responses} shows, GPT-4 was identified as human by 18 raters, surpassing the actual human expert, who was recognized by 16 raters.

 GPT-4 was frequently praised for being "well-structured," "empathetic," and "providing step-by-step guidance." The feedback highlighted that GPT-4's responses "bridge the gap between complex legal information and layperson accessibility," but it was also noted that there is a need to "reduce verbosity" and avoid "overly technical language" to prevent overwhelming users.  Claude 3, which also performed well quantitatively, received more mixed qualitative feedback. While its responses were described as "clear," "empathetic," and providing "practical advice," some evaluators noted an inconsistency, mentioning that the responses "lacked the depth and specific procedural details" needed for more concrete guidance. For ChatGPT 3.5, while the responses were noted to be "reasonably structured," they often "lacked depth" and were described as "generic," failing to provide sufficient specificity, such as detailing "legal notice contents." This gap between general information and specific legal guidance was seen as a detractor to both "perceived quality" and effectiveness in addressing user queries.

Human-generated responses, despite scoring lower quantitatively, were praised in qualitative feedback for their "empathy" and "personalization," which felt more "genuine and grounded in practical reality." However, their lower scores might stem from a more concise style that, while empathetic, lacked "exhaustive detail" seen in the top-performing models. This discrepancy suggests a gap in how human responses are measured compared to AI ones, particularly when human responses prioritize empathy and real-world relevance over sheer detail.
This dichotomy reveals an interesting dynamic: while AI models perform exceptionally well on objective criteria, human feedback underscores the value of qualitative elements such as empathy, contextual adaptation, and specificity that are less quantifiable but critically important in legal advisory contexts. Nonetheless, the results show that LLM chatbots can perform relatively well in offering tailored legal advice to clients. 

\subsection{Legal Research}

Legal research is the process of identifying and retrieving the legal information necessary to support legal decision-making, argumentation, and drafting. It involves finding and analysing statutes, regulations, case law, legal precedents, and secondary sources such as legal commentaries and journals. Previous studies have found that AI’s performance in research tasks varies considerably depending on the specificity of the task and the model’s training \citep{Tu2023}. While some custom LLM techniques have demonstrated promising results in finding relevant cases to given case-facts \citep{Izzidien2024,Ostling2024,Shu2024}, there have been no studies on the general capabilities of LLM chatbots for retrieving legal information. 

For the legal research task, respondents were required to evaluate the capabilities of Large Language Models (LLMs) against human experts in performing a core legal function: finding relevant case law for a specific legal issue. The task involved providing a list of pertinent cases in the scenario involving a consumer grievance where a client reported finding insects in a beverage at a restaurant, leading to physical and emotional distress. Both LLMs and the human—were tasked with identifying at least five appropriate cases from Indian judicial authorities, including the Supreme Court, the High Courts, or the National Consumer Disputes Redressal Commission (NCDRC). 

\begin{figure}[h]
  \centering
  \includegraphics[width=\linewidth]{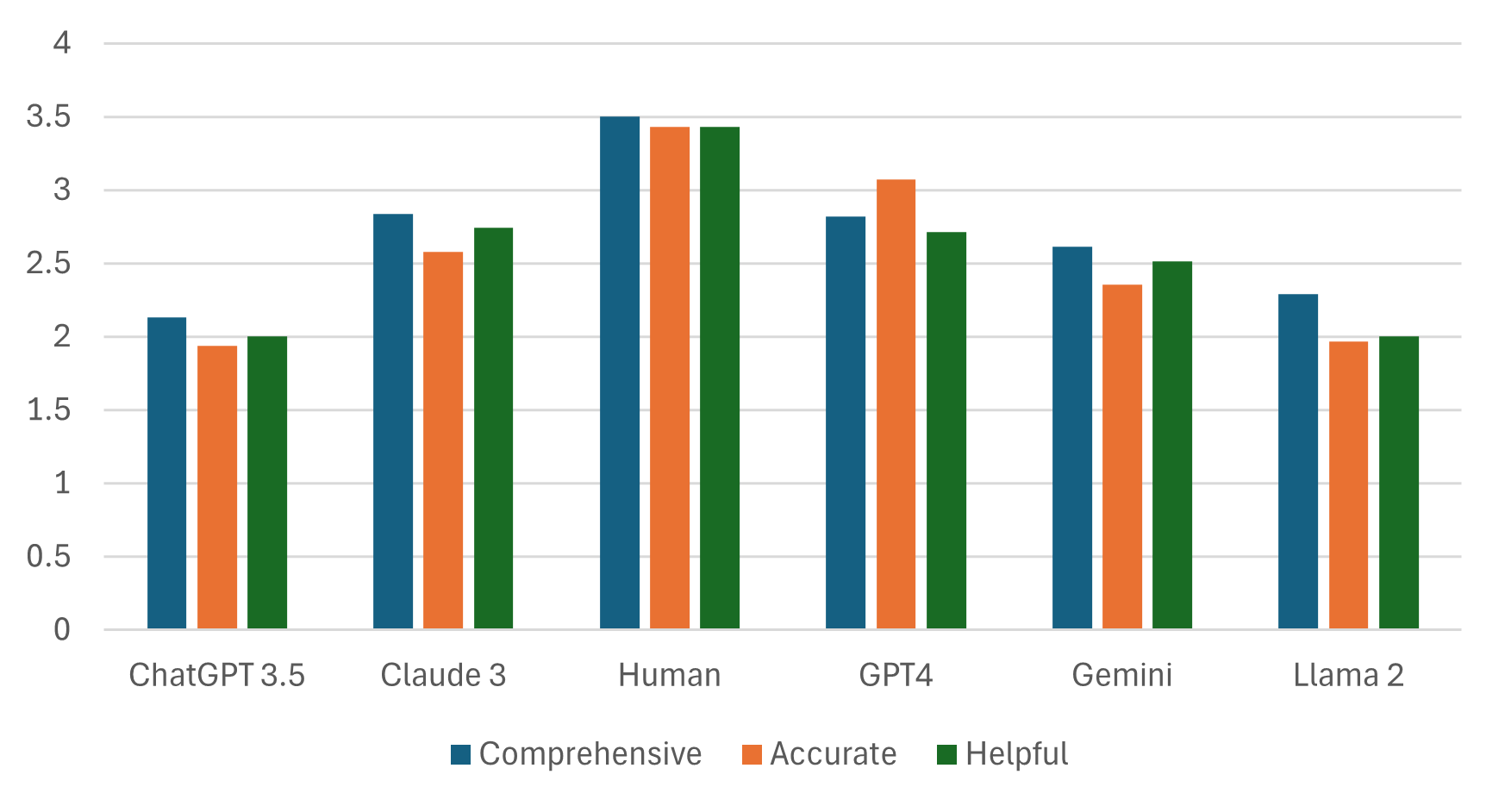}
  \caption{Bar chart of mean evaluation score of outputs for the legal research task.}
  \label{fig:research}
\end{figure}

Figure \ref{fig:research} presents the results of the legal research task. As can be seen, human respondents consistently outperformed all AI models in comprehensiveness, accuracy, and helpfulness, underscoring the critical limitations of current AI in legal research. Human experts scored the highest in all three categories. In contrast, every AI model, from GPT-4 to Claude 3, showed significant shortcomings, particularly with their tendency to hallucinate cases—fabricating case names, facts, or citations—which makes them unreliable for high-stakes legal research tasks. 
 
While GPT-4 and Claude 3 were occasionally praised for structuring legal arguments well and providing summaries that seemed coherent, their responses were consistently undermined by hallucinations. The outputs of GPT-4, for instance, were described as “convenient” for referencing but were criticised for "mentioning fake cases" such as "Neelkanth Venkatesh," which, after thorough searches, were found not to exist. The less sophisticated AI models, such as GPT-3.5 and LLaMA 2, were even more evidently flawed. Respondents noted these models' persistent issues with generating “fake cases” and lacking an understanding of “why relevant facts are necessary.” These hallucinations not only made the AI outputs unreliable but also unsuitable for any professional legal setting where accuracy is essential. In fact, a manual check indicated that \textit{all} AI generated cases were either hallucinated or were inapplicable to the facts at hand. 

 In contrast, human respondents were overwhelmingly more accurate and were often identified correctly as human-generated. The strength of human responses lies in their accuracy and the absence of fabricated cases. These responses included relevant facts and ratios, clearly citing real cases that could be directly applied to the legal issue at hand. Respondents noted that humans “mention both the relevant facts and the ratio in one paragraph” and “provide complete citations”. 

One possible reason for these results may be that many LLMs are trained predominantly on Western legal texts, resulting in limited or no exposure to Indian case law. This scarcity means that references to Indian legal decisions are either few and far between or entirely absent, forcing the models to rely on more familiar jurisdictions and increasing the likelihood of inaccuracies. Additionally, the context-specific nature of Indian legal cases requires an understanding of local statutes and judicial interpretations, which LLMs may not consistently possess. This may make it difficult for these models to provide relevant and reliable citations for any given legal issue.
The results therefore underscore that while AI models can mimic the structure of legal research to some extent, their propensity to hallucinate case law remains a significant barrier. Human respondents’ avoid these critical errors and provide precise, relevant, and reliable legal research. As it stands, AI models cannot yet replace the credibility and accuracy of human research, especially in contexts where credibility and correctness are non-negotiable.

\subsection{Legal Reasoning}

Legal reasoning is the process of "thinking like a lawyer," involving the logical analysis and application of legal principles to specific facts to reach a well-supported conclusion. In legal reasoning tasks, studies illustrate that LLMs can perform at levels close to humans in specific contexts, such as extracting case outcomes or applying rules. Most tests focus on Legal Judgement Prediction (LJP) which requires LLMs to correctly predict the decision in a case given a set of facts \citep{Wu2023}. Several studies have shown that fine-tuned LLM systems can mimic human decision-making \citep{Cao2024}. One study, found that GPT-4’s accuracy in extracting case outcomes was comparable to that of humans, demonstrating the potential of LLMs to handle large volumes of legal texts and derive meaningful insights \citep{Ostling2024}. Similarly, another comprehensive evaluation found that some LLMS could perform rule application, rule recall, rule conclusions, and interpretation with reasonable accuracy, depending on the nature of the task and the quality of the input prompts \citep{Guha2023}. 
 
\begin{figure}[h]
  \centering
  \includegraphics[width=\linewidth]{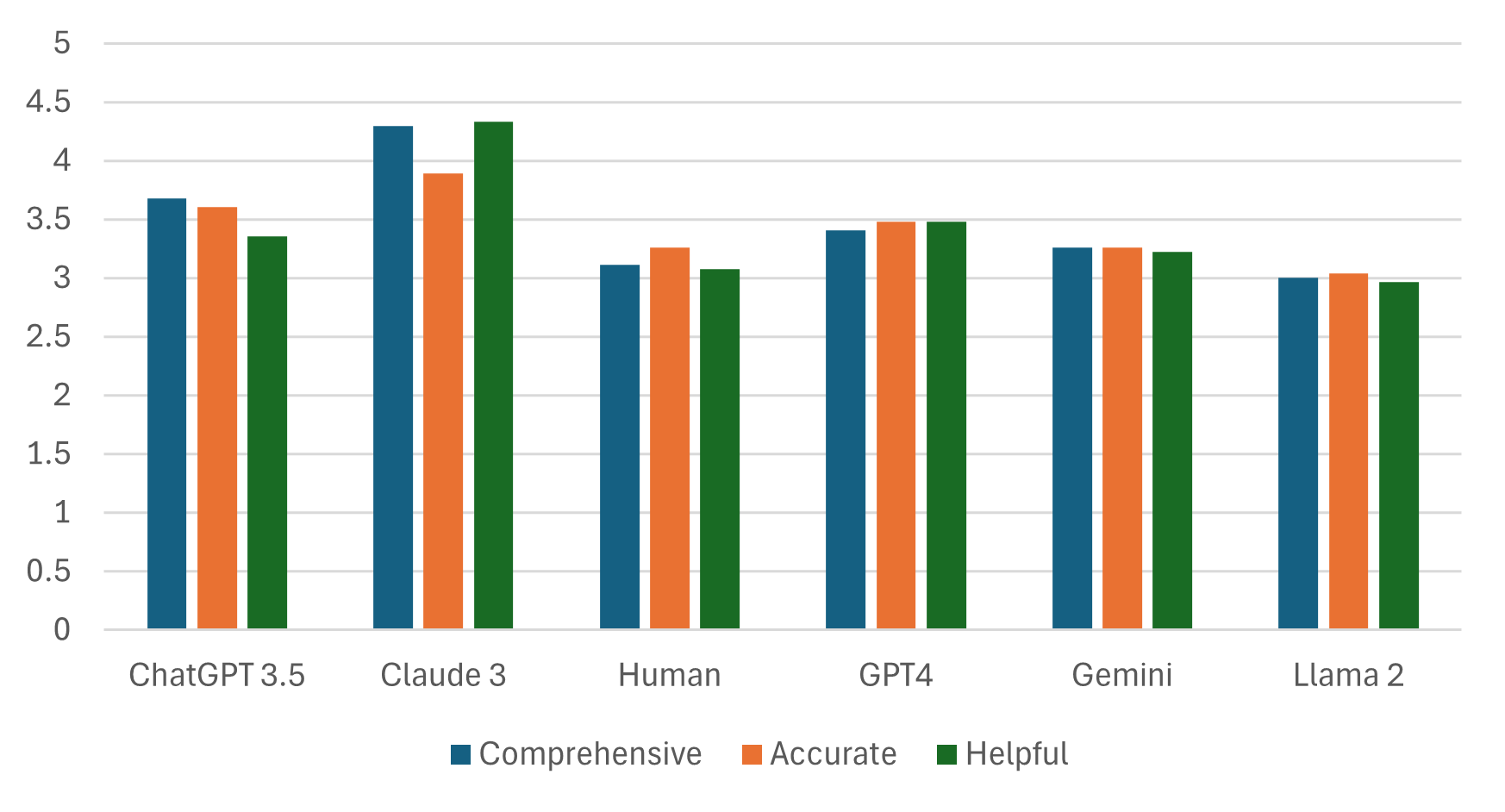}
  \caption{Bar chart of mean evaluation score of outputs for the legal reasoning task.}
  \label{fig:judge}
\end{figure}

In the legal reasoning task, respondents were required to evaluate the capability of Large Language Models (LLMs) in drafting legal judgments in comparison to human experts. This exercise simulates a scenario where participants, acting as judges, are tasked with drafting a judgment for a hypothetical case involving a consumer grievance filed under the Consumer Protection Act, 2019, in India. The provided case details revolve around a complaint by an individual against a hotel for serving a beverage with insects, resulting in distress and a demand for compensation. The participants were expected to analyse the facts presented in the complaint, the counterarguments by the opposite party (the hotel) and apply relevant legal principles to craft a comprehensive judgment. The drafting of the judgment required respondents to integrate findings, legal reasoning, conclusions, and orders based on the facts and applicable law. The participants were assessed on their ability to articulate a coherent and legally sound decision, capturing the core elements of judicial reasoning, including the analysis of liability, the application of legal doctrines such as "volenti non fit injuria" (the principle that one who consents to an act cannot claim injury), and the proper remedy if any.

Figure \ref{judge} presents the results. As can be seen, chatbots outperform human respondents in key dimensions of drafting legal judgments. Notably, Claude 3 achieved the highest scores across all categories, surpassing both Gemini and GPT-4. Llama 2 lagged slightly behind other chatbots but still exceeded human performance in comprehensiveness and accuracy. In terms of respondents identifying the chatbot-generated judgments as human, Claude 3 again stood out, indicating a high level of sophistication and human-like reasoning in its responses  (\ref{tab:human_responses}). 

Qualitative feedback supports these findings, as respondents praised Claude 3 for its "well-reasoned judgment" and "accurately recalls case law and applies it correctly." Additionally, Claude 3 was often identified as human due to its "structured and detailed analysis" and "professional language," which contributed to its high helpfulness score. This indicates that Claude 3’s ability to integrate legal principles and articulate coherent judgments effectively meets the expectations of respondents, making it stand out among both AI models and human experts. 

In contrast, respondents felt that while humans provided "clear and logical conclusions," they sometimes lacked the depth and structured analysis seen in chatbot responses. For instance, one respondent noted that human judgments "deal with each issue separately" but also mentioned that they could "be more detailed in legal analysis." This discrepancy highlights that although humans bring nuanced reasoning and contextual understanding, they may not always present their judgments with the same level of structured comprehensiveness and factual accuracy as advanced language models like Claude 3.

These results show that LLMs can apply legal reasoning, at least as effectively as humans and provide structured, accurate, and professionally articulated judgments. They particularly underscore the potential of LLMs to support judicial functions, which face similar time and resource constraints as the human expert. 

\section{Discussion}

The results give us quite a detailed picture of the capabilities of LLMs. Across all tasks, except legal research, LLMs performed comparably with the human expert. Across the evaluated tasks, GPT-4 and Claude 3 consistently emerged as top performers, demonstrating strengths in comprehensiveness, accuracy, and helpfulness. Notably, Claude 3 showcased exceptional proficiency in Legal Drafting and Legal Reasoning, often matching or exceeding human experts in producing detailed, accurate, and persuasive legal documents and judgments. The open-source LLaMA 2, in particular, performed poorly across many tasks. Outputs by human experts, while generally reliable and precise, were occasionally outperformed by the most advanced AI models in structured tasks like issue spotting and legal reasoning. However, human-generated outputs demonstrated superior capabilities in tasks requiring deep contextual understanding, empathy, and strategic legal thinking, such as legal drafting and legal research.

At the core of the observed performance disparities lies the fundamental distinction between language processing and domain-specific knowledge application. LLMs like GPT-4 and Claude 3 demonstrated exceptional capabilities in tasks that involve extracting, summarizing, and synthesizing existing information. This proficiency can be attributed to the inherent design of LLMs, which are optimized for understanding and manipulating natural language through extensive training on diverse textual data. Consequently, these models are adept at handling language-intensive tasks that require the aggregation and presentation of information in a coherent and accessible manner.

However, this linguistic strength does not uniformly translate to all aspects of legal practice. In tasks that necessitate specific legal knowledge, precise factual accuracy, and the application of nuanced legal principles, LLMs exhibited notable deficiencies. The legal research task, for example, exposed significant limitations in AI performance, particularly the propensity for hallucinations—instances where models generate plausible-sounding but factually incorrect information. The tendency to fabricate case laws or misrepresent legal doctrines highlights a critical vulnerability of LLMs in high-stakes legal environments where accuracy and reliability are paramount. This shortfall is primarily due to the models' reliance on probabilistic language generation rather than a robust, verifiable understanding of legal statutes and precedents.

The disparity in performance across tasks can also be explained by the nature of training data and the specificity of legal knowledge required. While LLMs like GPT-4 and Claude 3 are trained on vast corpora that include general legal texts, they may lack the depth and contextual awareness necessary for specialized legal tasks, especially those that are in highly localised legal systems such as consumer law in India. Legal reasoning often involves intricate interpretation of statutes, case law analysis, and the application of legal principles to unique factual scenarios—processes that demand more than surface-level understanding of the language of law. Human experts, through education and experience, develop a sophisticated grasp of these elements, enabling them to apply social knowledge to navigate complex legal landscapes. This inherent advantage of human cognition underscores the current limitations of AI in fully replicating the depth of legal expertise.

The qualitative feedback further elucidates the nuanced performance of LLMs compared to human experts. Advanced models like Claude 3 were frequently mistaken for human-generated content due to their high precision and structured output, indicating their ability to emulate human-like reasoning and presentation. However, evaluators also noted instances where these models lacked the strategic variability and emotional nuance characteristic of human legal professionals. These attributes are difficult to quantify and replicate through AI, highlighting the complementary strengths of human-AI collaboration rather than AI as a standalone replacement.

What I also found while building the survey is that prompt engineering is very important.  The outputs generated by LLMs were the result of robust prompt engineering. Well-defined prompts that closely mirror real-world legal scenarios enable LLMs to generate more accurate and relevant outputs. In contrast, vague or broad instructions can lead to generic or erroneous responses, exacerbating issues like hallucinations and reducing the reliability of AI-generated content. This dependency on prompt quality suggests that while LLMs possess the potential to assist in legal tasks, their effectiveness is contingent upon the precision of task definitions and the contextual relevance of their training data. 

\section{Conclusion}

The comparative performance of LLMs and human experts in legal tasks reveals a landscape of both promise and caution. Advanced AI models like GPT-4 and Claude 3 exhibited significant capabilities in language processing and structured task execution, showing a high degree of proficiency in certain legal tasks. This suggests significant potential for AI to augment legal professionals by automating routine analyses and document drafting. This augmentation can enhance efficiency, reduce workload, and allow human lawyers to focus on more complex, strategic, and client-centric aspects of their practice. 

Yet, the potential for AI to replace lawyers is not absolute. For example, tasks requiring strategic thinking, empathy, and ethical judgment, such as advocacy, negotiation, and client counselling, are beyond the current capabilities of AI, as these involve interpersonal skills and deep contextual understanding that AI cannot replicate. Additionally, AI technology cannot perform tasks that require physical presence and activity, such as representing clients in court or engaging in face-to-face negotiations. The path forward lies in leveraging the complementary strengths of AI and human professionals, fostering collaborative environments where technology augments rather than replaces human judgment, and continuing to advance AI capabilities to better meet the intricate demands of legal practice.

This study has some important limitations. First, the research used a convenience sample of 50 law students, which is relatively small and may not represent the broader legal profession. There is a need for further empirical research on this topic, with larger and more diverse sample pools. Second, I could only assess human performance against a single junior lawyer. Comparing with additional human outputs of different types may help grade the level of human expertise that AI reaches.  Third, AI technology changes rapidly over time. The models evaluated were tested in April, and since then, more advanced versions, like ChatGPT 4o, LLaMA 3.1, and Claude 3.5, have been released. Preliminary assessments suggest these newer models, such as o-1, designed for reasoning tasks, may outperform those tested in this study. Thus, the findings offer only a snapshot of a fast-evolving field, underscoring the need for ongoing evaluation. 

What I have also not touched on in this paper is also that the use of AI in legal settings requires lawyers to navigate ethical complexities that AI systems are ill-equipped to handle \citep{Rogers2019}. The legal profession is governed by a framework of ethical standards that necessitate human judgment and accountability, which AI lacks. This is particularly relevant in sensitive areas such as family law and criminal defence, where the stakes are high, and the consequences of legal decisions can profoundly impact individuals' lives. Lawyering often requires a deep understanding of clients' emotional states and experiences, a skill set that AI cannot adequately provide. For AI systems to become more prevalent in legal practice, concerns about accountability, data privacy, and algorithmic bias will also have to be addressed.

Nonetheless, I believe that this study offers important insight, highlighting both the potential and limitations of AI in specific legal contexts. As AI technology advances, continuous research will be essential to monitor progress, address shortcomings, and explore new applications in the legal profession. On the AI side, enhancing the factual accuracy of LLMs, developing robust verification mechanisms to prevent hallucinations, and refining models to better understand and apply complex legal doctrines are critical areas for future development. On the human side, training legal professionals to effectively utilize AI tools, understand their capabilities and limitations, and maintain critical oversight will be essential in harnessing the benefits of AI.

The results of this study suggest a future where lawyers collaborate with AI, leveraging its strengths to enhance efficiency and accuracy in their practice rather than being entirely replaced by it. The integration of AI into the legal profession is likely to enhance certain aspects of legal work, but it will not replace the fundamental role of lawyers.

\bibliographystyle{apalike}
\bibliography{references}

\end{document}